Research Paper

# Semantics-aware next-best-view planning for efficient search and detection of task-relevant plant parts

Akshay K. Burusa *, Joost Scholten, Xin Wang, David Rapado-Rincón, Eldert J. van Henten, Gert Kootstra

*Agricultural Biosystems Engineering, Wageningen University & Research, Wageningen, the Netherlands*

ARTICLE INFO

ABSTRACT

Searching and detecting the task-relevant parts of plants is important to automate harvesting and de-leafing of tomato plants using robots. This is challenging due to high levels of occlusion in tomato plants. Active vision is a promising approach in which the robot strategically plans its camera viewpoints to overcome occlusion and improve perception accuracy. However, current active-vision algorithms cannot differentiate between relevant and irrelevant plant parts and spend time on perceiving irrelevant plant parts. This work proposed a semantics-aware active-vision strategy that uses semantic information to identify the relevant plant parts and prioritise them during view planning. The proposed strategy was evaluated on the task of searching and detecting the relevant plant parts using simulation and real-world experiments. In simulation experiments, the semantics-aware strategy proposed could search and detect 81.8% of the relevant plant parts using nine viewpoints. It was significantly faster and detected more plant parts than predefined, random, and volumetric active-vision strategies that do not use semantic information. The strategy proposed was also robust to uncertainty in plant and plant-part positions, plant complexity, and different viewpoint-sampling strategies. In real-world experiments, the semantics-aware strategy could search and detect 82.7% of the relevant plant parts using seven viewpoints, under complex greenhouse conditions with natural variation and occlusion, natural illumination, sensor noise, and uncertainty in camera poses. The results of this work clearly indicate the advantage of using semantics-aware active vision for targeted perception of plant parts and its applicability in the real world. It can significantly improve the efficiency of automated harvesting and de-leafing in tomato crop production.

Nomenclature

| | |
|---|---|
| 2D | Two dimensional |
| 3D | Three dimensional |
| 6D | Six dimensional |
| DoF | Degrees of freedom |
| NBV | Next-best view |
| OOIs | Objects of interest |
| OPTICS | Ordering points to identify the clustering structure |
| PCO | Percentage of correctly-detected objects |
| RGB-D | Red Green Blue colour and depth |
| ROS | Robotic operating system |

## 1. Introduction

Tomatoes are among the most consumed vegetables globally (Beed et al., 2021) with demand growing rapidly due to increasing population. However, the current rate of tomato production cannot satisfy the future demand and efforts to improve production are limited by labour shortages. The current labour force is aging, as younger generations are not interested in the physically demanding and monotonous work in the tomato industry (Rigg et al., 2020). To improve production and meet the rising demand, there is a strong need for automation, where robots can fill-in for humans and carry out the labour-intensive tasks in tomato greenhouses, such as selective harvesting and de-leafing (Van Henten, 2004).

To automate harvesting and de-leafing of tomato plants, robots must accurately perceive the relevant plant parts. For harvesting, it is relevant to perceive ripe tomatoes and peduncles (part that connects tomatoes to the stem). Similarly, for de-leafing, it is relevant to perceive the petioles (part that connects leaves to the stem). These plant parts are shown in Fig. 1. With advancements in computer vision, the 2D detection of tomatoes, peduncles, and petioles can be achieved using cameras and artificial neural networks (Koirala et al., 2019). However, to estimate the cutting point for harvesting and de-leafing, detection on a 2D image alone is not sufficient. We need to gather more information about the plant parts and estimate their cutting point in 3D. Hence, searching and sufficiently perceiving all the relevant parts of a tomato plant remains to be a challenging problem, primarily due to occlusion by other parts (Kootstra et al., 2021), leading to incomplete or ambiguous information.

* Corresponding author.
*E-mail address:* akshaykumar.burusa@wur.nl (A.K. Burusa).

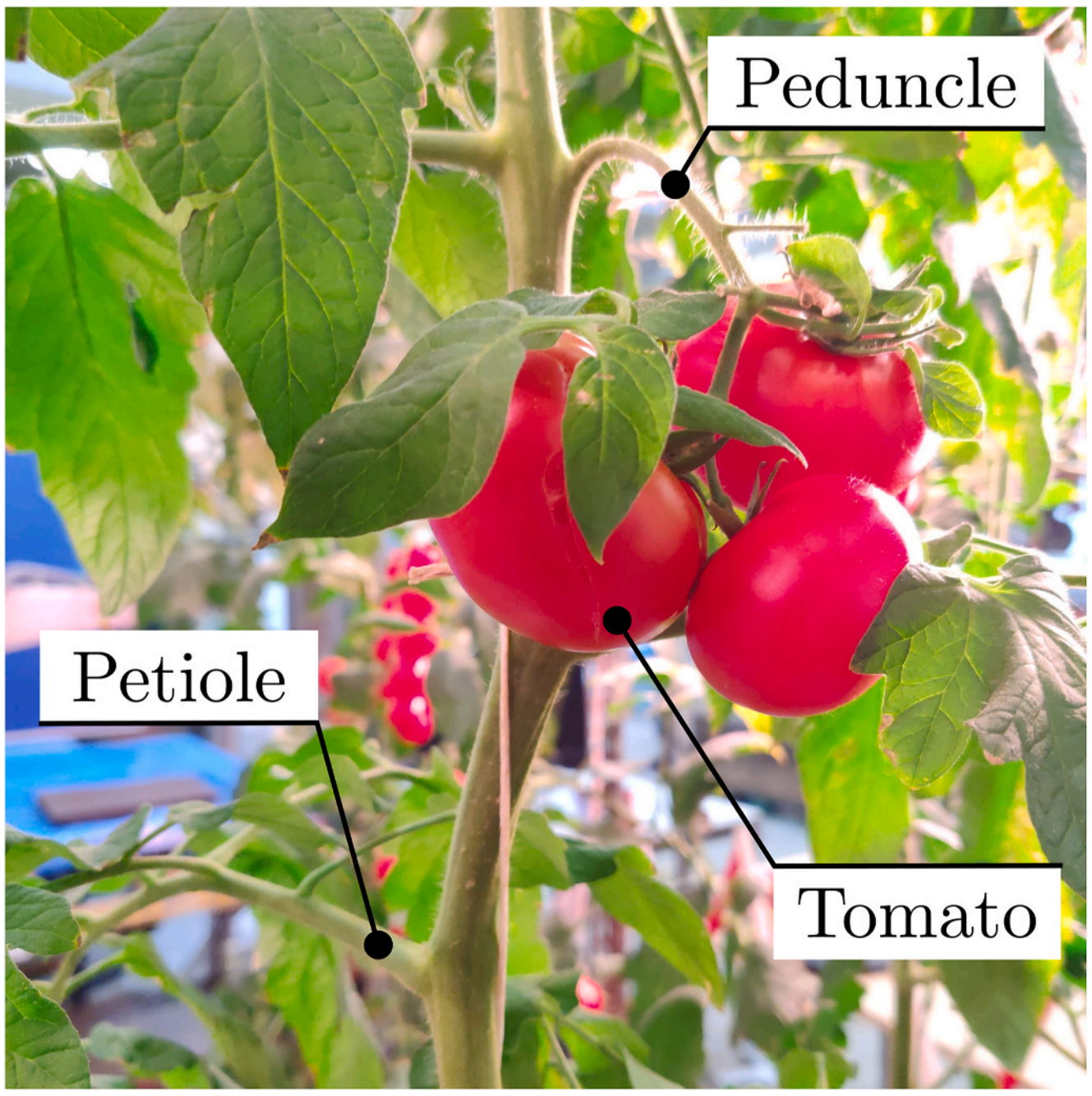

**Fig. 1.** Efficiently searching and finding the relevant plant-parts, such as tomatoes, peduncles, and petioles, is essential for automated harvesting and de-leafing in tomato greenhouses.

Many approaches to occlusion handling are proposed in the robotics literature and applied in agro-food environments. With advancements in deep learning, a popular approach is to train a neural network to detect objects despite occlusions or to complete the shape of the occluded parts. However, such approaches rely on training data and might not extend well to situations that are out-of-distribution. Also, these approaches cannot be applied to greenhouse tasks that require careful inspection of the plant parts, such as disease detection. In such situations, the best way to guarantee certainty of perception is to physically move the camera and directly observe the occluded plant part. Hence, one way to overcome occlusion is to explore the plant using multiple viewpoints, as acquiring multiple measurements can help fill any missing information or resolve ambiguities (Hemming et al., 2014; Van Henten et al., 2002). However, these multiple viewpoints must be chosen efficiently. In current practice, passive multi-view approaches are used, in which a camera is moved in predefined patterns (e.g. zig-zag motions). Such approaches often suffer from sub-optimal view selection, i.e., they are either time consuming due to too many viewpoints or fail to detect all plant parts due to too few viewpoints.

Active vision (Bircher et al., 2016; Daudelin & Campbell, 2017; Isler et al., 2016; Schmid et al., 2020; Zeng et al., 2020) offers a better approach for viewpoint planning, in which the camera viewpoints are deliberately planned to maximise novel information about the plant based on the information perceived so far. By maximising novel information, active vision prioritises viewpoints that observe new or occluded parts of the plant. Hence, it can explore the plant faster and more efficiently compared to passive multi-view approaches (Burusa et al., 2024; Marangoz et al., 2022; Menon et al., 2023; Zaenker et al., 2021, 2023). However, most active-vision algorithms are designed to explore unknown space using occupancy information, which is ideal for exploring the whole plant. The occupancy information cannot distinguish between different plant parts and hence is not suitable for targeted perception of specific objects-of-interest (OOIs), such as the relevant plant parts. To address this problem, active-vision algorithms must be made aware of the OOIs. Adding semantic information about objects, such as class labels, has been shown to improve targeted perception (Kay et al., 2021) and scene understanding (Zheng et al., 2019) in other applications. Recent works in greenhouse robotics (Marangoz et al., 2022; Menon et al., 2023; Zaenker et al., 2021, 2023) made a binary distinction between fruits and the rest of the plant and showed improvement in the estimation of the shape, size, volume, and position of fruits. However, these works did not use the semantic information directly for planning viewpoints and were only evaluated in simulation. Hence, their quantitative performance under real-world conditions is unknown.

In this paper, a semantics-aware active-vision strategy that considers the class labels and confidence scores of the OOIs when planning viewpoints is proposed. It was hypothesised that using such semantic information will lead to faster perception of all the relevant plant parts. This hypothesis was evaluated using simulation and real-world experiments. In simulation, 3D mesh models of tomato plants with varying growth stages and structural complexity, and uncertainty in plant and OOI positions were used. With real-world experiments, the complexities of the real world, such as natural variation and occlusion in tomato plants, natural illumination, sensor noise, and uncertainty in camera poses were introduced. Using these setups, systematic and repeatable experiments were performed to study the advantage of adding semantic information to active vision. The performance of the proposed semantics-aware active vision method was compared with an active vision strategy without semantic information, predefined strategies, and a random strategy. Also, to the best of our knowledge, this work was the first to evaluate the performance of a semantics-aware active vision approach using a robotic setup in a real-world greenhouse environment.

## 2. Problem description

Given a tomato plant, placed within a bounded 3D space $V \subset R^3$, the task is to explore the plant using a robot with an RGB-D camera and detect all the OOIs. The robot does not know the exact position of the plant, except that the plant is somewhere in front of it. Hence, the position of the plant base $p^{pb} \in V$ contains some uncertainty. The robot starts from an initial camera viewpoint $\xi_0$ and needs to find a set of sequential viewpoints that will lead to the detection of the OOIs. Here, a camera viewpoint $\xi$ can have 6 degrees-of-freedom (DoF), with a 3D position and 3D orientation. Since the structure of the plant is not known *a priori*, the robot needs to simultaneously explore the plant to find new OOIs and improve the perception of previously detected OOIs. This task of visual search for OOIs was formulated as an online next-best-view (NBV) planning problem within the active-vision paradigm. The goal of NBV planning is to determine the next-best camera viewpoint $\xi_{best}$ that will help perceive novel information about the OOIs. It was assumed that by maximising the novel information about the OOIs per viewpoint, the shortest sequence of viewpoints that can detect most of the OOIs with high accuracy can be found. The problem is terminated when all the OOIs are detected or when a maximum number of viewing actions $a_{max}$ is reached.

## 3. Semantics-aware next-best-view planning

A semantics-aware NBV planning algorithm was proposed to address the problem described above. The main loop of the planner is illustrated in Fig. 2. The algorithm consisted of three major modules – (i) a sensing module to detect the OOIs, (ii) a 3D scene representation module to merge information about the OOIs across multiple viewpoints, and (iii) a view-planning module to determine the next-best camera viewpoint.

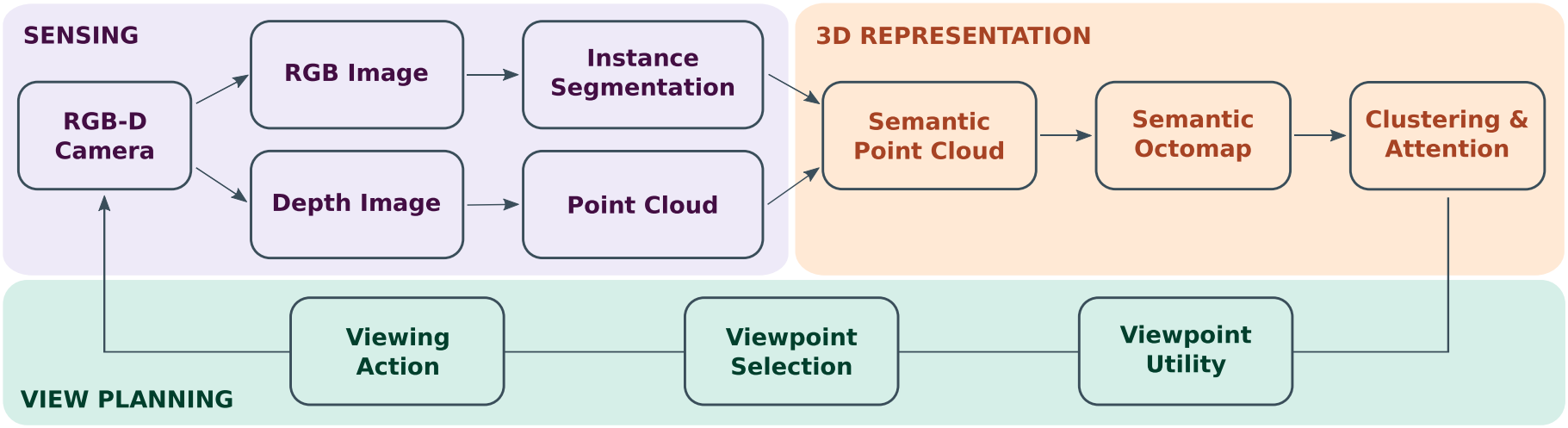

**Fig. 2.** The pipeline of the Semantic NBV planning algorithm consists of three major modules – (i) a sensing module to detect OOIs (Section 3.1), (ii) a 3D scene representation module to analyse and track the OOIs over multiple viewpoints (Sections 3.2 to 3.5), and (iii) a view-planning module to determine the next-best camera viewpoint (Sections 3.6, 3.7).

### *3.1. Sensing*

The first step of the algorithm was to accurately detect the OOIs. For this, a convolutional neural network, Mask R–CNN (He et al., 2017) was used, since such techniques have proven to be effective for the perception of plant parts (Jiang & Li, 2020). For simulation experiments, the network was fine tuned to detect tomatoes, peduncles, and petioles using a custom dataset of 90 RGB images, of dimension 960 × 540 pixels, collected from the simulated environment. The dataset was split into 72 images for training and 18 images for validation. For real-world experiments, we focused only on the detection of peduncles and petioles, as these parts are crucial for harvesting and de-leafing and are more challenging to perceive than tomatoes due to occlusion. Mask R–CNN was fine-tuned to detect peduncles and petioles using a custom dataset of 746 RGB images collected in a tomato greenhouse on seven different occasions. A total of 658 peduncles and 1478 petioles were annotated. 526 images were used for training and 220 images for validation. The dataset is available at https://github.com/akshaykburusa/tomato_greenhouse_dataset. The object detection module took an RGB image as input and generated a segmented image with masks that separated the OOIs from the background (Fig. 3a).

### *3.2. Semantic point cloud*

The segmented image was combined with the corresponding depth image to generate a semantic point cloud (Fig. 3b). Each point $i$ in the point cloud contained the spatial position of the OOIs in Cartesian space and the semantic class label $c_s(i)$ and confidence score $p_s(i)$ about what object the point belonged to. The class label and confidence score were defined as,

$$c_s(i) \in \{-1 = \text{background}, 0 = \text{peduncle}, 1 = \text{petiole}, 2 = \text{tomato}\}, \quad (1)$$

$$p_s(i) \in [0, 1]. \quad (2)$$

The confidence score $p_s(i)$ was interpreted as the probability of the point $i$ belonging to the detected class label $c_s(i)$. Hence, $p_s(i) = 0$ meant that the point certainly did not belong to $c_s(i)$, whereas $p_s(i) = 1$ meant that the point certainly belonged to $c_s(i)$. The values in between implied uncertainty regarding the class label, with maximum uncertainty at $p_s(i) = 0.5$. If a point did not contain any semantic information, a default value of $c_s(i) = -1$ and $p_s(i) = 0.5$ were assigned, which meant that the point belonged to the background and had maximum uncertainty regarding its class label.

### *3.3. Semantic OctoMap*

The semantic point cloud captured the spatial and semantic information of the plant only in the current viewpoint. To keep track of the detected OOIs across multiple viewpoints, a 3D scene representation of the plant which could be updated over time was needed. So, the observed plant and its surrounding scene were represented as a probabilistic occupancy map called an OctoMap (Hornung et al., 2013). The OctoMap divided the 3D space into grid cells called voxels, which were efficiently stored as hierarchical nodes in an Octree. A resolution of 0.003 m for a single voxel was used. Each voxel was marked as empty or occupied based on whether the voxel was observed by the camera. Voxels with an occupancy probability of zero ($p_o = 0$) were considered *empty* and those with occupancy probability of one ($p_o = 1$) were considered *occupied*. A value between 0 and 1 implied uncertainty regarding the occupancy, with a maximum uncertainty at $p_o = 0.5$. The regions that were not observed by the camera yet did not contain any occupancy information and were considered *unknown* with maximum uncertainty ($p_o = 0.5$). Such information about the occupancy probabilities of the voxels is commonly referred to as *volumetric information*. The volumetric information is useful in identifying the uncertain and unexplored regions of the plant and can help guide an NBV planner towards them. However, it cannot distinguish between different plant parts and hence not suitable for targeted perception of the OOIs.

To be able to explore the OOIs, the OctoMap was extended to include *semantic information* in addition to the volumetric information (Fig. 3c), similar to Xuan and David (2018). That is, each voxel $x$ in the OctoMap consisted of a semantic class label $c_s(x)$ and a confidence score $p_s(x)$ which were obtained and inserted from the semantic point cloud in the previous step (Section 3.2). The class labels identified the OOIs and distinguished them from the rest of the plant, while the confidence scores indicated how certain we were about the OOI identities. Hence, the semantic OctoMap could help guide an NBV planner to search for OOIs and further explore them until a degree of certainty was achieved.

The semantic information in the OctoMap was updated iteratively when the camera was moved to a new viewpoint and novel information was observed. If the voxel had no prior semantic information, the observed values were directly assigned to it. Otherwise, the semantic information was updated using the method of max-fusion (Xuan & David, 2018). The new class label and confidence score from the semantic point cloud were merged with the previous values in the voxel as follows: (i) if the new class label was the same as the previous, then the class label remained unchanged and the confidence score was computed as the average of the new and previous scores; or (ii) if the new class label was different from the previous, the semantic label with the higher confidence score was chosen and that confidence score was reduced by 10% as a penalty for label mismatch. This penalty increased the uncertainty of the voxel, which in turn could guide the viewpoint planner to take another look at the voxel and resolve the label mismatch. The max-fusion method was used as it was simple, fast, and memory-efficient as it only stored one class label and confidence score per voxel. The pseudo code for the max-fusion method is shown in Algorithm 1.

**Algorithm 1.** Max fusion for merging semantic information

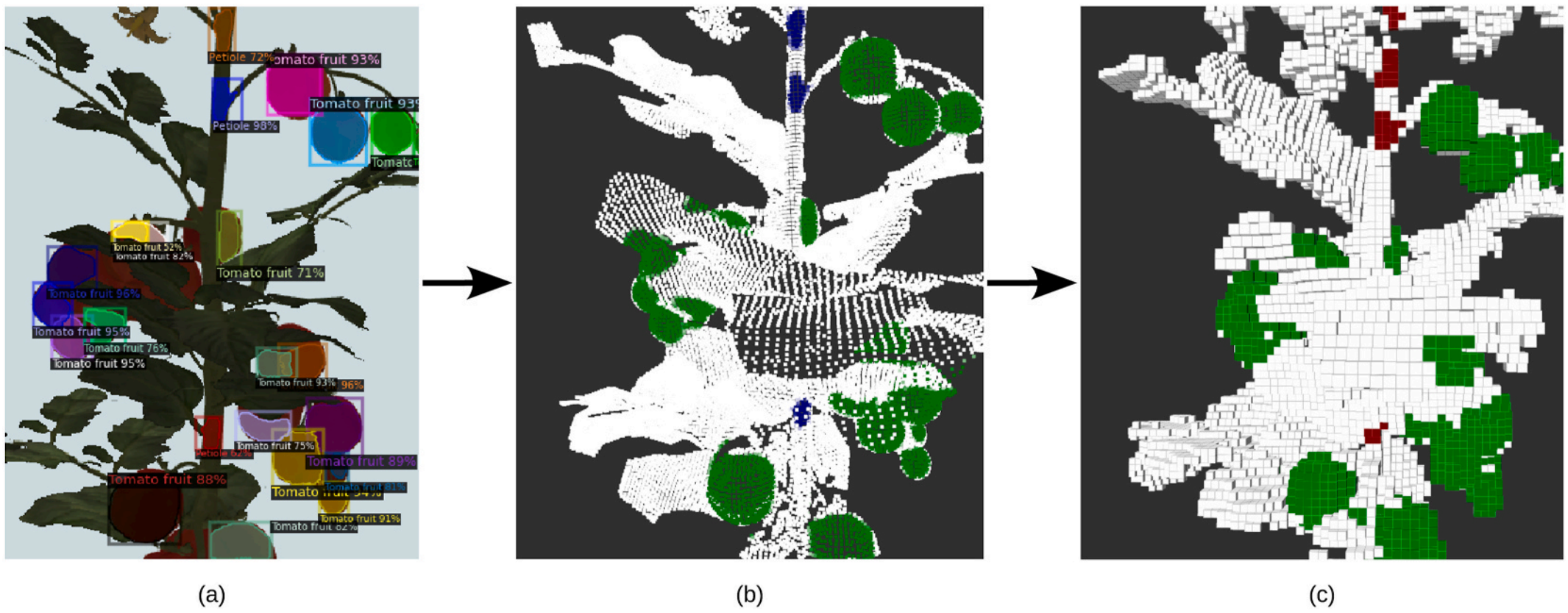

**Fig. 3.** Generating the semantic OctoMap had the following steps: (a) The objects-of-interest (OOIs) were detected using Mask R–CNN to form a segmented image (Section 3.1). (b) The segmented image was combined with the corresponding depth image to produce a semantic point cloud (Section 3.2). (c) The semantic point cloud was then inserted into the OctoMap, where each voxel stores the class label and confidence score, along with the occupancy probability (Section 3.3). Also, a clustering operation was performed on the semantic OctoMap to identify the individual OOIs at an object level in 3D space (Section 3.4).

**Algorithm 1** Max fusion for merging semantic information

$\mathrm{sem}_1, \mathrm{sem}_2$ ▷ Previous and new semantics
$\mathrm{sem}_f$ ▷ Updated semantics
$c_s, p_s$ ▷ Class label and confidence score
**function** MAX FUSION($\mathrm{sem}_1, \mathrm{sem}_2$)
  **if** $\mathrm{sem}_1.c_s == \mathrm{sem}_2.c_s$ **then**
    $\mathrm{sem}_f.c_s \leftarrow \mathrm{sem}_2.c_s$
    $\mathrm{sem}_f.p_s \leftarrow (\mathrm{sem}_1.p_s + \mathrm{sem}_2.p_s)/2$
  **else**
    $\mathrm{sem}_f.c_s \leftarrow \arg\max_{c_s}(\mathrm{sem}_1.p_s, \mathrm{sem}_2.p_s)$
    $\mathrm{sem}_f.p_s \leftarrow 0.9 \times \max(\mathrm{sem}_1.p_s, \mathrm{sem}_2.p_s)$
  **end if**
  **return** $\mathrm{sem}_f$
**end function**

### 3.4. *Clustering*

The semantic OctoMap represented the OOIs in a discretised form at the voxel level. However, an object-level representation was required to locate the OOIs in 3D space. This was achieved by clustering the voxels that belonged to the same object. An algorithm proposed by Ankerst et al. (1999) called *Ordering Points To Identify the Clustering Structure* (OPTICS) was used. Each cluster was identified by three criteria – (i) all voxels in the cluster must belong to an OOI, (ii) the total number of voxels in a cluster must be at least 20, which was empirically chosen based on the clustering performance, and (iii) the maximum distance between any two voxels within the cluster is 0.03m for simulation experiments and 0.06m for real-world experiments. These distances were based on the size of the OOIs. With this clustering step, an object-level representation of the OOIs was obtained. The number of detected OOIs was given by the number of clusters, and the positions of the OOIs $p^{\mathrm{ooi}} \in V$ were given by the centre of the clusters. The clustering was repeated at every iteration of the Semantic NBV planning pipeline. Hence, as the semantic OctoMap was updated over multiple viewpoints (Section 3.3), the OOIs were also updated.

### 3.5. *Attention mechanism*

An object-level representation of OOIs allowed the NBV planner to attend to regions where the OOIs were located and improve their accuracy. So, similar to the previous work of Burusa et al. (2024), an attention mechanism[1] was used to guide the NBV planner to explore and perceive the OOIs. The attention regions were marked by defining bounding boxes around the estimated position of the main stem and the detected positions of the OOIs, as shown in Fig. 4. The main-stem bounding box guided the planner to explore and identify new OOIs and the OOI bounding boxes guided the planner to perceive the previously detected OOIs more accurately. Zaenker et al. (2021) defined two separate sampling strategies to obtain these behaviours, a target-based sampling to generate viewpoints that observed the previously detected OOIs and a frontier-based sampling to generate viewpoints that searched for new OOIs. In this paper, these behaviors were obtained using a single strategy based on the bounding boxes. The bounding boxes were defined using the centre position of the box, the dimensions of the sides, and the orientation with respect to the global frame. The regions of attention for the main stem and the OOIs were defined as follows.

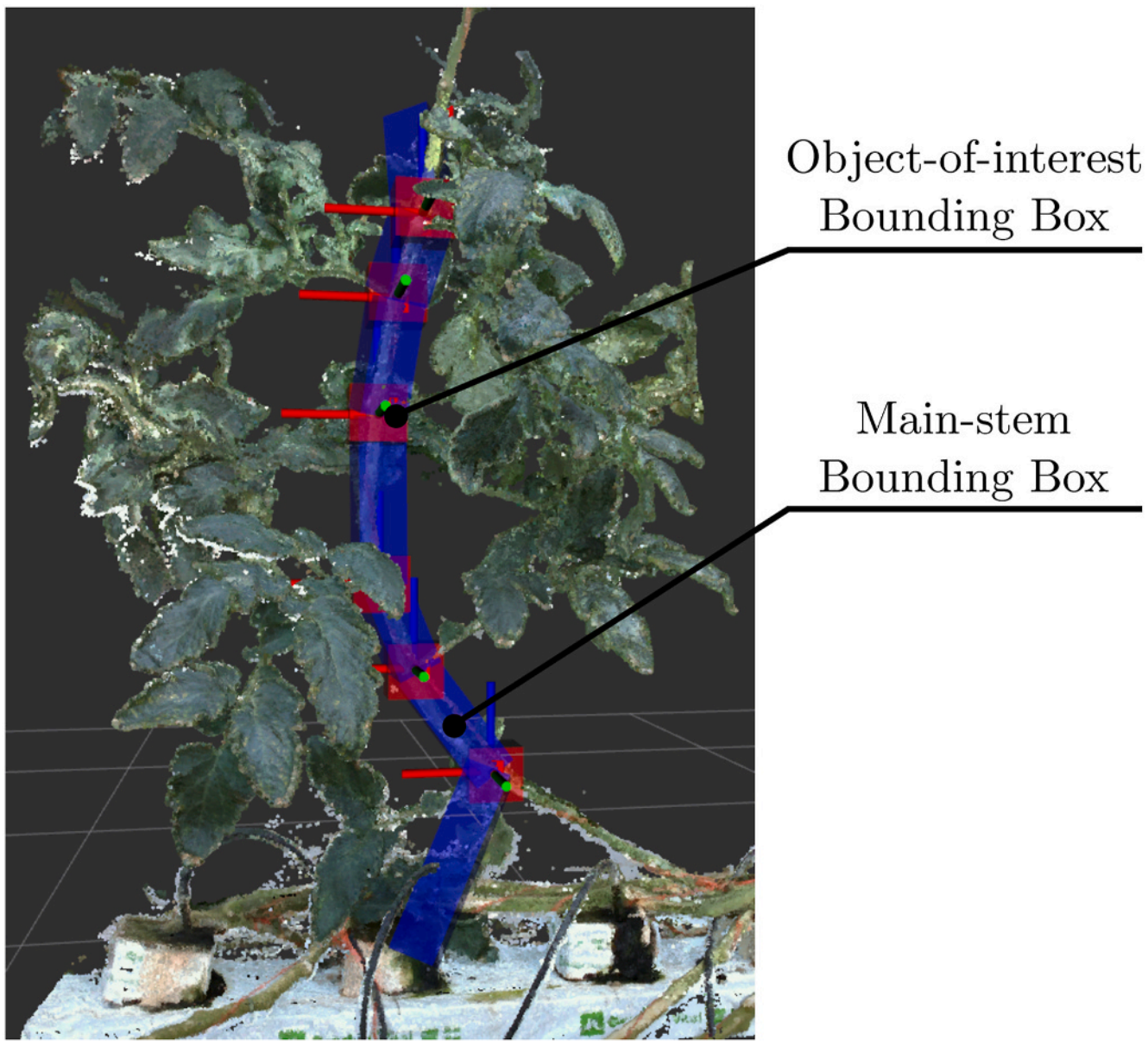

**Fig. 4.** An attention mechanism was used to guide the Semantic NBV planner. A region-of-interest was defined along the estimated position of the main stem (blue bounding box) to guide the planner to explore new OOIs. Multiple regions-of-interest were defined around the detected OOIs (red bounding boxes) to guide the planner to perceive them more accurately. The axes represent the 3D positions of the OOIs.

[1] Not to be confused with the attention mechanism in Transformer architectures in deep learning.

#### 3.5.1. Attention along the estimated main stem

The main stem was not detected, but its position was estimated based on the volumetric and semantic information gathered so far, as explained below. Bounding boxes were defined around this estimated position of the main stem. The position estimation of the main stem depended on how much information we had about the plant. In particular, there were four different cases. (i) When there was no information about the plant, attention along the main stem was not defined. In this case, the NBV planner planned viewpoints without an attention mechanism. (ii) Once the plant was viewed at least from one viewpoint, the position of the main-stem centre was estimated as the centre of the visible region of the plant. An axis-aligned bounding box of height 0.7 m in simulation and 1.2 m in the real-world experiments was defined at this estimated position to guide the NBV planner to explore the plant further. Here, axis-aligned bounding box means that the edges of the box were parallel to the global coordinate frame. (iii) Once some tomato clusters were detected, the main-stem centre was estimated as the centre of the tomato clusters instead, as this was a better estimate of the stem location than the centre of all visible regions of the plant. (iv) Finally, when at least one petiole or peduncle cluster was detected, a set of oriented bounding boxes were defined between consecutive petioles or peduncles, as shown in Fig. 4. Two boxes at the top and bottom of the bounded space $V$ (which is expected to contain the plant) were also included, to ensure that the NBV planner also explored the top and bottom regions of the stem. Hence, the attention on the main stem improved iteratively as more information was obtained from multiple viewpoints. The breadth and width of the bounding boxes were always set to 0.05m.

#### 3.5.2. Attention on the detected OOIs

An OOI bounding box was defined only when an OOI cluster was detected (Section 3.4). The position of the OOI $p^{ooi}$ was used to define the centre of the OOI bounding box. The size of the bounding box was predefined as a cube with size 0.03m in simulation and 0.06m in the real-world experiments, according to the size of the objects. The bounding boxes were axis-aligned (Fig. 4). These bounding boxes automatically adapted to any changes in the OOI clusters such as a shift in position or were completely removed if the cluster was not detected any more.

### 3.6. Viewpoint utility

The selection of the next-best viewpoint was guided by a metric called *expected information gain*, which estimated the amount of novel information that could be obtained when a camera was moved to a given viewpoint. For the Semantic NBV planner, the information gain was defined using the semantic information in the OctoMap.

#### 3.6.1. Expected semantic information gain

The semantic information gain is a measure of how much novel semantic information can be gained by exploring a 3D volume. The semantic information $I_{\text{sem}}$ expected to be gained by observing a voxel $x$ was defined as its entropy,

$$I_{\text{sem}}(x) = -p_s(x)\log_2(p_s(x)) - (1-p_s(x))\log_2(1-p_s(x)), \tag{3}$$

where $p_s(x)$ is the confidence score that the voxel $x$ belonged to a certain object. The confidence score was obtained from the semantic OctoMap as discussed in Section 3.3. The information gain has a range between 0 and 1. Intuitively, it is high for voxels with high uncertainty (i.e. $p_s(x)$ close to 0.5) and low for voxels with low uncertainty (i.e. $p_s(x)$ close to 0 or 1). Hence, it is a measure of uncertainty.

To guide the Semantic NBV planner to search for OOIs, it was designed to prioritise viewpoints that maximised the expected semantic information gain. The expected semantic information gain $G_{\text{sem}}(\xi)$ for a viewpoint $\xi$ was computed as the sum of the semantic information of all voxels that were expected to be visible from $\xi$, that is,

$$G_{\text{sem}}(\xi) = \sum_{x\in(\mathscr{X}_\xi \cap \mathscr{B})} I_{\text{sem}}(x), \tag{4}$$

where $\mathscr{X}_\xi$ was the set of all voxels that were expected to be visible from viewpoint $\xi$ and $\mathscr{B}$ was the set of all voxels that were within the attention regions (defined in Section 3.5). $\mathscr{X}_\xi$ was determined by the process of ray-tracing, in which a set of rays were cast from viewpoint $\xi$ evenly across the camera's view frustum. Each ray traversed until it hit an occupied voxel or reached a maximum distance. All voxels that the rays intersected, were expected to be visible if the camera was moved to viewpoint $\xi$ and hence were added to the set $\mathscr{X}_\xi$.

The Semantic NBV planner used only the semantic information and discarded the volumetric information. Such a formulation however did not limit the planner from understanding occupied and unoccupied regions of space, since the semantic information inherently contains occupancy information to some extent. For example, any voxel that contains a semantic class label is, by definition, occupied. However, the Semantic NBV planner gave no importance to how certain we were about the occupancy and focused primarily on how certain we were about the semantic labels.

#### 3.6.2. Total viewpoint utility

In addition to the expected semantic information gain, the distance from the current viewpoint was also considered when selecting the next viewpoint since there was a time cost for executing the motion. The time cost was approximated as the Euclidean distance between the current viewpoint and the next potential viewpoint. So, the final view utility $U$ was the expected semantic information gain $G_{\text{sem}}(\xi)$ scaled by the Euclidean distance cost $d$,

$$U_{\text{sem}} = G_{\text{sem}}(\xi) \times e^{-d}, \tag{5}$$

where the exponential to the negative distance cost prioritised the viewpoints that were very close to the current viewpoint.

Now, the objective of NBV planner was to find a viewpoint that maximised the viewpoint utility, which would lead to efficient exploration and detection of the OOIs.

### 3.7. Viewpoint selection

To determine the next-best viewpoint, a set of candidates $\mathscr{V}$ was sampled at each step within a planar-surface constraint, defined in Section 4.2.1. These candidates were potential viewpoints where the camera could be moved next. For each candidate, the view utility was computed using Equation (5), and the viewpoint with the maximum utility was chosen as the next-best viewpoint,

$$\xi_{\text{best}} = \arg\max_{\xi\in\mathscr{V}} U_{sem}(\xi). \tag{6}$$

Sampling the candidates is an important step for view selection. If all the candidates were poor choices, we would end up with an inefficient sequence of viewpoints despite maximising the information gain at each step. Hence, a pseudo-random sampling strategy was used, similar to the work of Burusa et al. (2024), which evenly distributed the candidates across the planar surface. This strategy tried to ensure that there were at least a few candidates whose selection would lead to efficient exploration of the OOIs, although this was not guaranteed. The number of sampled candidates was 27 for the simulation experiments and 60 for the real-world experiments.

## 4. Experiments

### 4.1. Existing methods for comparison

The performance of the Semantic NBV planner was compared with a Volumetric NBV planner (Burusa et al., 2024), two predefined planners and a random planner as baselines.

#### 4.1.1. Volumetric NBV planner

The Volumetric NBV planner is a common approach in NBV planning, which uses the expected gain in volumetric information to determine the next-best viewpoint. Many state-of-the-art approaches in view planning are built upon volumetric or occupancy information (Zeng et al., 2020). The Volumetric NBV planner tends to explore all uncertain regions of the plant, which is only suitable for whole-plant perception. Similar to semantic information gain, the volumetric information gain is a measure of how much novel volumetric information can be gained by exploring a 3D volume. The volumetric information $I_{\text{vol}}$ expected to be gained by observing a single voxel $x$ was defined as,

$$I_{\text{vol}}(x) = -p_o(x)\log_2(p_o(x)) - (1 - p_o(x))\log_2(1 - p_o(x)), \quad (7)$$

where $p_o(x)$ is the probability of voxel $x$ being occupied. The expected volumetric information gain $G_{\text{vol}}(\xi)$, the viewpoint utility $U_{\text{vol}}$, and the next-best viewpoint $\xi_{\text{best}}$ were computed similar to Equations (4)–(6) respectively. Also an adaptive attention mechanism for the Volumetric NBV planner to focus on the main stem (similar to Section 3.5.1) was provided. The Volumetric NBV planner was similar to the Semantic NBV planner, expect that it used $G_{\text{vol}}(\xi)$ instead of $G_{\text{sem}}(\xi)$ and did not use OOI bounding boxes.

#### 4.1.2. Predefined planners

Two predefined planners with a zig-zag pattern were used, which is a common practice in greenhouse robotics to detect the OOIs. Since there was a large uncertainty in plant position along the y-axis in the simulation experiments, a single predefined planner that could perceive the plant well across the whole uncertainty range ($y = \pm 0.3$ m) could not be designed. So, two planners with different ranges of motion were used: (i) A Predefined Narrow planner with a narrow range making an angle of 18° to the centre of the plant, and (ii) a Predefined Wide planner with a wider range making an angle of 54° (Fig. 5). The orientations of the viewpoints were fixed as xyzw: $(0.0, 0.0, 0.0, 1.0)^T$ defined as a quaternion, such that camera was facing the plant. The predefined planners scanned the plants from the bottom to top.

#### 4.1.3. Random planner

A random planner that chose one viewpoint at random from the set of candidate viewpoints V that were sampled at each step of the Semantic NBV planner (Section 3.7) was used. This planner was used a baseline to confirm that the Semantic NBV planner was more efficient than choosing a viewpoint at random.

### 4.2. Simulation experiments

We studied how efficiently could the Semantic NBV planner detect the OOIs compared to the Volumetric NBV, Predefined Wide, Predefined Narrow, and Random planners. The simulation setup (Section 4.2.1) contained uncertainty in plant and OOI positions, high level of occlusions from plant parts, and a planar-surface constraint for sampling the candidate viewpoints. The evaluation criteria for the comparison are discussed in Section 4.2.3.

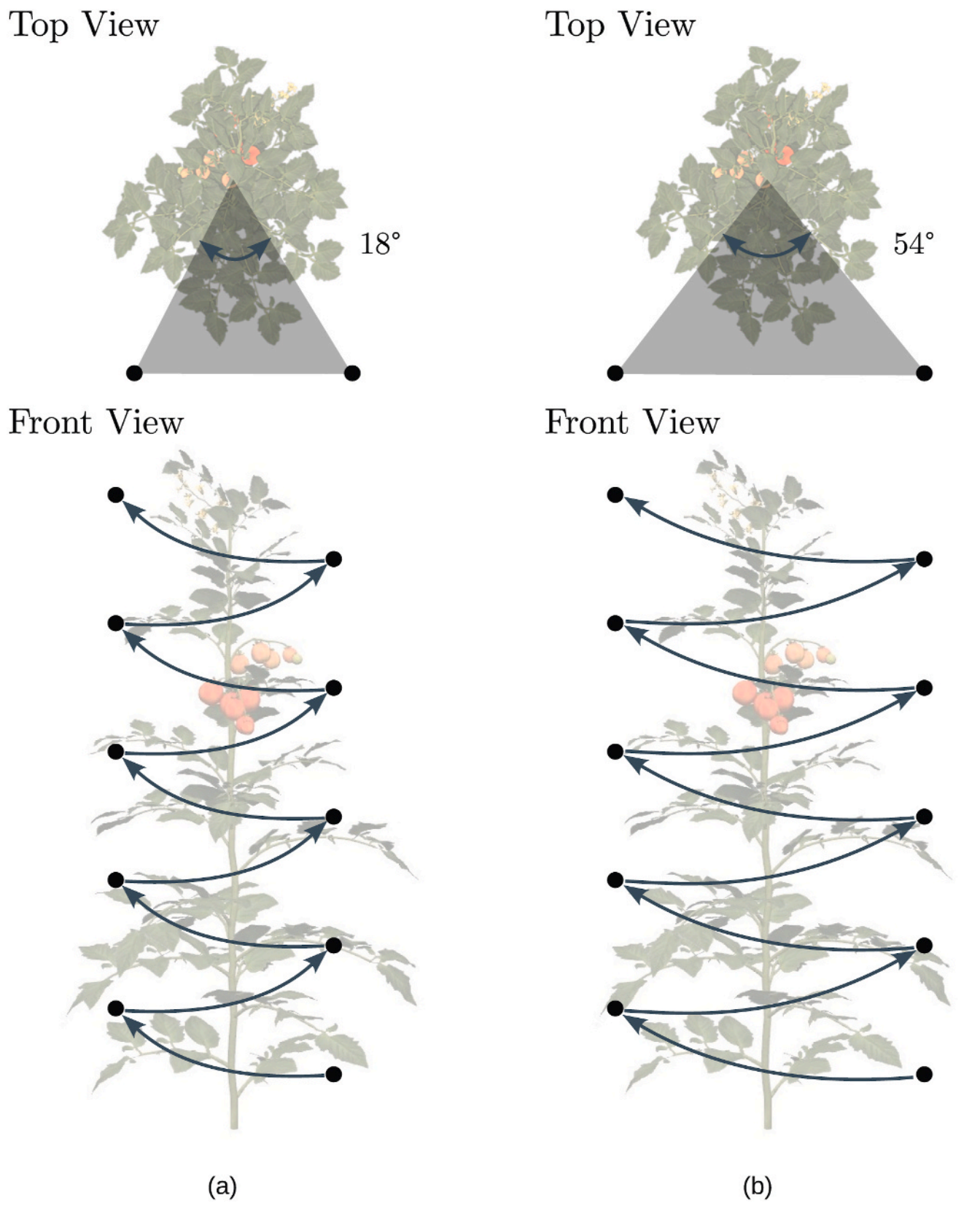

**Fig. 5.** The performance of the Semantic NBV planner was compared with a Volumetric NBV planner, two predefined planners and a random planner. The predefined planners had ten viewpoints and differed in their range of motion: (a) Predefined Narrow planner and (b) Predefined Wide planner.

#### 4.2.1. Simulation setup

A simulated environment with *Gazebo*[2] was used, as illustrated in Fig. 6. The robotic setup consisted of a 6 DoF robotic arm (ABB's IRB

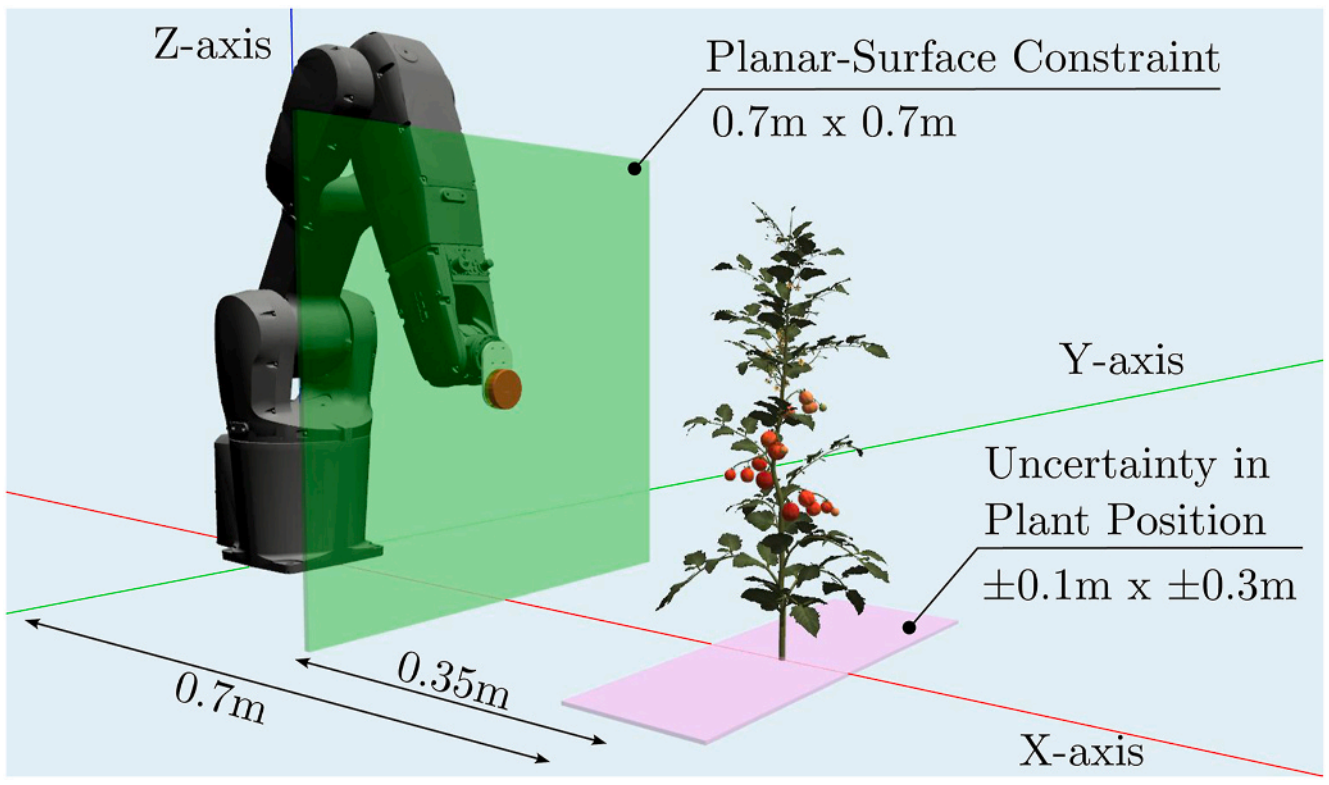

**Fig. 6.** Schematic diagram of the experimental setup in simulation, with an ABB IRB 1200 robotic arm and Realsense L515 camera. The plant was positioned in front of the robot with uncertainty (shown in pink) to test the robustness of the viewpoint planners. The robot was allowed to choose viewpoints on a planar-surface constraint (shown in green) with 4 DoF.

[2] http://gazebosim.org/.

1200) with an RGB-D camera (Intel Realsense L515) attached to its end-effector. Eight 3D mesh models[3] of tomato plants of varying growth stages and structural complexity were used, which the robot explored to detect the OOIs (illustrated in Fig. 7). A tomato plant was placed in front of the robot at $p^{\mathrm{pb}} = (0.7 \pm 0.1\mathrm{m}, 0.0 \pm 0.3\mathrm{m}, 0.8\mathrm{m})$. Uncertainty was introduced in the plant-base position by uniformly sampling the x and y positions in the range $0.7 \pm 0.1$ m and $0.0 \pm 0.3$ m respectively. The experiments were repeated for 12 different rotations of the plant along the z-axis at intervals of 30° each, leading to a total of 96 experiments.

A planar-surface constraint was used for view planning, i.e., the robot could choose viewpoints only within the bounds of the plane. The planar surface was of size $0.7 \times 0.7$ m$^2$, with its centre at a distance of 0.35m from the plant and at a height of 0.8m. In addition to positioning the camera on the planar surface, the viewpoints could also make a pan and tilt of $0 \pm 15°$, which were uniformly sampled. Hence, the viewpoints had 4 DoF. All the viewpoint planners started from the same initial viewpoint $\xi_0$ with position xyz: $(0.35, 0.22, 0.90)^T$ and orientation xyzw: $(0.0, 0.0, -0.26, 0.97)^T$. The robot was allowed to choose a maximum of $a_{\max} = 10$ consecutive viewing actions to explore the plant and identify as many OOIs as possible.

#### 4.2.2. Analysis of experimental and model parameters

We analysed the effect of the following experimental and model parameters on the planner performance:

**Impact of removing uncertainty in plant position:** In practice, the exact position of the tomato plants will not be known. Hence, the Semantic NBV planner should be able to handle such uncertainty. In the simulation experiments, the plant positions had an uncertainty of $\pm$ 0.1 m in the x-axis and $\pm$0.3 m in the y-axis. The impact of this uncertainty was analysed by comparing the performance of the planners to a case where the exact plant positions were assumed to be known.

**Impact of removing uncertainty in OOI positions:** The Semantic NBV planner must deal with uncertainty in the OOI positions too, which could be caused due to errors in detection. The Semantic NBV planner should still be able to estimate the OOI positions accurately and define appropriate bounding boxes for attention. The impact of OOI-position uncertainty was analysed by comparing the performance of the planners to a case where the exact OOI-positions were known.

**Impact of removing attention:** The attention mechanism proposed is crucial in improving the efficiency of OOI detection. Without an attention mechanism, the planner will consider semantic information from all voxels expected to be visible from a viewpoint. Hence, it might prioritise every part of the plant instead of the OOIs alone, which could make it inefficient. This hypothesis was tested by comparing the performance of the planners to a case where the main-stem bounding boxes and the OOI bounding boxes were completely removed.

**Impact of reducing plant complexity:** The Semantic NBV planner needs to detect the OOIs even when they are heavily occluded from the camera's view. The impact of occlusion was analysed by comparing the performance of the planners to a case where the plant complexity was reduced by removing around 50–60% of the leaflets, as shown in Fig. 7b.

**Sensitivity to sampling constraint:** The planar-surface constraint (Section 4.2.1) defined a region in which the planners could sample candidate viewpoints. The sensitivity of the planners to different sampling constraints was analysed to ensure that the results would still hold under different sampling strategies. The performance of the planners were compared under a planar-surface constraint (4 DoF) and a cylindrical-sector constraint (2 DoF), as used in Burusa et al. (2024). Under the cylindrical-sector constraint, a viewpoint could be positioned anywhere on the cylinder surface but was always oriented towards the centre of the cylinder. The cylinder had a height of 0.7m, radius of 0.4m and a sector angle of 90°. The plant positions were subjected to an uncertainty of $\pm$0.1 m in the x-axis and $\pm$0.3 m in the y-axis.

#### 4.2.3. Evaluation

The performance of the planners was evaluated by calculating the percentage of OOIs that were correctly detected out of all the OOIs in each plant. This metric is called the *percentage of correctly-detected objects* (PCO),

$$\mathrm{PCO} = \frac{\text{Correctly detected objects}}{\text{Total number of objects}} \times 100. \tag{8}$$

An object was considered correctly detected only when its semantic class was correctly identified and at least 50% of the object was accurately reconstructed by the planner. This was quantified using the metric of F1-score, similar to Burusa et al. (2024). The F1-score compares the ground-truth and reconstructed OOIs and estimates both the correctness and completeness of the reconstruction. It takes a value between 0 and 1, where a value of 1 implies a complete and correct reconstruction. The PCO metric is more rigorous than simply evaluating the 2D detection performance, as it requires the semantic class label, the position, and the shape of the object to be correctly perceived. Also, to accurately reconstruct 50% of the object despite occlusions, the object needs to be perceived correctly from multiple viewpoints.

A series of steps was followed to enable the computation of the F1-score for all the OOIs. First, the original 3D meshes of the tomato plants were converted to a point cloud and down sampled to the same resolution as the semantic OctoMap (i.e. 0.003m) using voxel-grid filtering. The semantic OctoMap was also converted to a point cloud, to make it comparable to the ground-truth. Only the occupied voxels were converted, i.e. voxels with $p_o(x) > 0.5$. Second, the true position of the OOIs were obtained manually by processing the 3D mesh of the plants in Blender.[4] For each object, the true position was calculated by averaging the positions of mesh vertices that belonged to the object, as shown in Fig. 8. Finally, from both the ground-truth and reconstructed point clouds, the points belonging to each OOI was extracted by using a bounding-box of size 0.03m and placing it at the true position of the OOI. The points within the bounding box were considered to belong to the OOI. These steps gave us the ground-truth and reconstructed point clouds for each OOI, which were then used to compute the F1-scores.

### 4.3. Real-world experiments

We performed real-world experiments with pre-recorded data collected from a tomato greenhouse at Unifarm, Wageningen University and Research, The Netherlands, to test if the Semantic NBV planner can perform accurately and efficiently under real-world complexities such as natural variation and occlusion, natural and varying illumination, sensor noise, and uncertainty in camera poses. The objective of the viewpoint planner was to search and detect the petiole and peduncles, as these are the plant parts that are relevant for de-leafing and harvesting tasks respectively. Hence, they were defined as the OOIs. The performance of the Semantic NBV planner was compared with Volumetric NBV, predefined, and random planners. All the viewpoint planners started from the same initial viewpoint $\xi_0$ for each plant. The robot was allowed to choose a maximum of $a_{\max} = 10$ consecutive viewing actions to explore the plant and identify as many OOIs as possible.

#### 4.3.1. Real-world setup and data collection

The real-world robot was similar to the simulation robot, consisting of a 6 DoF robotic arm (ABB's IRB 1200) with an RGB-D camera (Intel Realsense L515) attached to its end effector. The position of the camera with respect to the robotic arm was calibrated using eye-in-hand calibration. Robotic Operating System (ROS) was used as the middleware and a motion planning package for ROS, MoveIt!, was used for moving the camera to a desired viewpoint. The robotic setup was mounted on a cart that could move on the rails between rows of tomato plants in a greenhouse.

[3] https://www.cgtrader.com/3d-models/plant/other/xfrogplants-tomato.

[4] https://www.blender.org/.

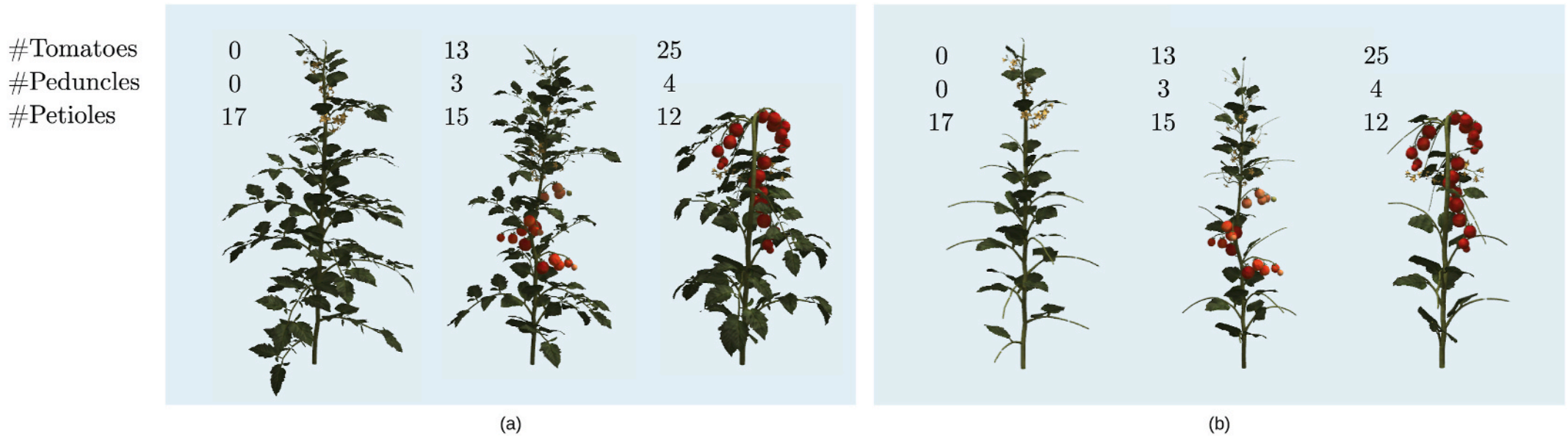

**Fig. 7.** Eight mesh models of tomato plants were used for the experiments. The plants varied in terms of growth stage, number of OOIs, height and complexity. Some examples of (a) the original plant models and (b) the same plant models with around 50–60% leaflets removed are shown.

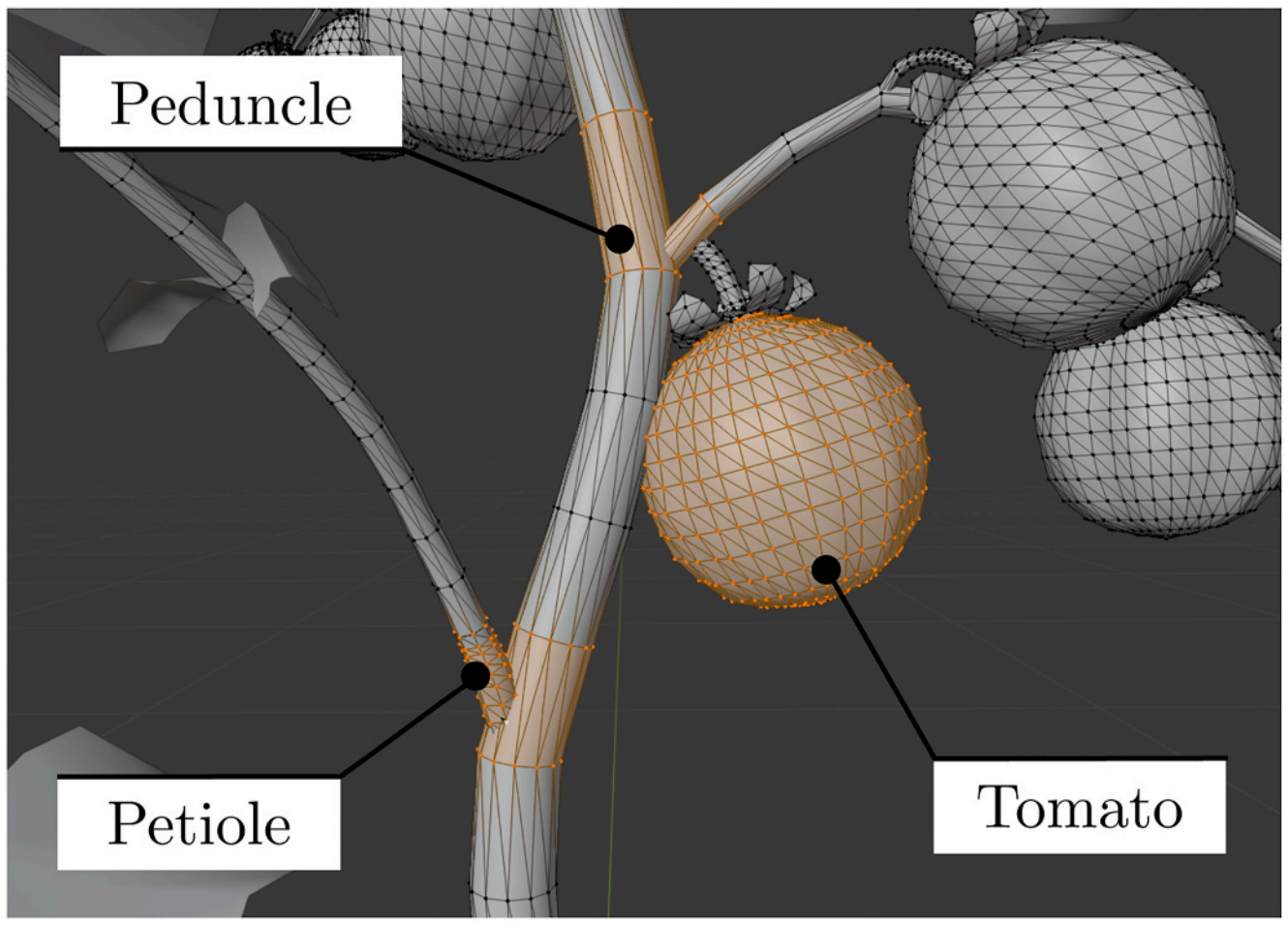

**Fig. 8.** The ground truth for evaluation was obtained by processing the original 3D mesh of tomato plants in Blender. The ground-truth positions of the objects-of-interest (OOIs) were estimated by averaging the positions of vertices that belonged to the object. Examples of a tomato, petiole, and peduncle with their vertices highlighted (in orange) is shown.

A dataset of ten tomato plants was collected. Each tomato plant had only one main stem, that is, any side shoots were removed by the growers. There were multiple tomato plants in a row, such the plant parts of a tomato plant were easily occluded by the leaves of adjacent plants. The complexity of the plants in the greenhouse can be seen in Fig. 9. For the dataset, the colour and depth images with a resolution of $960 \times 540$ pixels, the corresponding structured point cloud, and the 6D camera pose with respect to the robot's frame were recorded. The viewpoints were predefined along a grid of 15 equally-spaced rows and 10 equally-spaced columns on a planar surface of height 0.6m and width 0.4m, with the centre of the plane at a distance of 0.5m from the plant and at a height of 0.75m from the robot's base frame. At each point on the grid, data was collected from five viewpoints with pan and tilt angles of $[(-15^\circ, 0^\circ), (15^\circ, 0^\circ), (0^\circ, 0^\circ), (0^\circ, 15^\circ), (0^\circ, -15^\circ)]$. Hence, the viewpoints had 4 DoF and made a total of 750 viewpoints for each plant, as shown in Fig. 9b.

This pre-recorded dataset allowed us to repeat experiments and compare the performances of different planners. The objective of the planners was to determine the shortest sequence of viewpoints from the 750 candidates that could efficiently find and perceive the petioles and peduncles. This was a discretised case of viewpoint planning in which the planners only had a subset of viewpoint to choose from. This was done for the sake of feasibility and repeatability of the experiments. When applied in practice, the planners can sample viewpoints uniformly from a continuous space that is reachable by the robot, as done in the simulation experiments. For each plant, the experiment was repeated 12 times, leading to a total of 120 experiments with ten real-world plants.

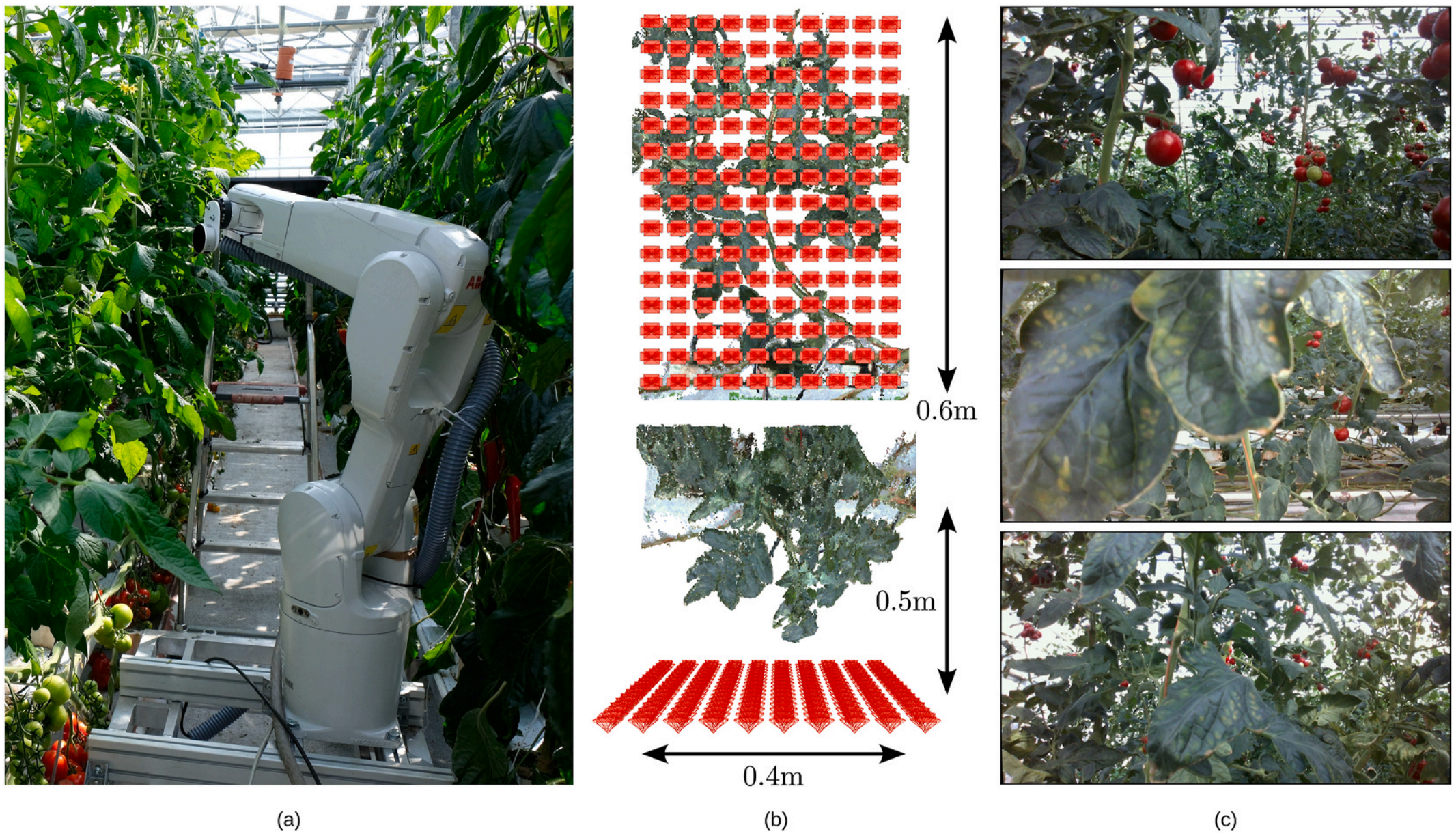

**Fig. 9.** (a) The real-world setup with a 6 DoF ABB IRB 1200 robotic arm and a Realsense L515 RGB-D camera. (b) Front and top views of the 750 viewpoints on a planar surface used for collecting data of tomato plants. (c) Examples of images of tomato plants collected by the robot, showing the level of occlusion.

#### *4.3.2. Ground truth and evaluation*

The true mesh models of real-world plants were not available, unlike the simulation experiments. Hence, the reconstruction from the planners was compared with COLMAP (Schonberger & Frahm, 2016), a structure-from-motion-based reconstruction algorithm. The colour images from all the 750 viewpoints for each plant, along with the camera poses, were provided to COLMAP to generate a 3D point cloud of the plant. On this point cloud, the 3D positions of the plant nodes were manually annotated by selecting all the points that belonged to each node and taking an average of their 3D positions. This COLMAP point cloud and the annotated node positions (as shown in Fig. 10) were used as ground truth in the evaluation.

The performance of the planners was evaluated similar to the simulation experiments, by computing the percentage of correctly-detected objects (PCO). An object was considered as correctly detected when it had an F1-score of at least 62.5%. This F1-score threshold was increased compared to the simulation experiments because the ground-truth point cloud was incomplete. The ground-truth point cloud was generated from a set of 750 predefined viewpoints which observed the plant only from one side. Based on empirical observations, it was assumed that the ground-truth point cloud represented less than 80% of the true OOI. So, to evaluate if 50% of the true OOI was observed, the F1-score threshold had to be increased to 62.5% of the ground-truth point cloud. The F1-score compared the OOI points in the reconstructed point cloud from the semantic OctoMap and the ground-truth point cloud from COLMAP. A similar procedure described in Section 4.2.3 was followed, except for two details. First, the point clouds were down sampled to a resolution of two-voxels distance (i.e. 0.006m) instead of one-voxel distance, to account for the fact that the COLMAP point cloud was not a perfect ground truth and had some reconstruction error. Second, the size of the bounding box used to extract points belonging to the OOIs was increased to 0.04m, since the OOIs in the real-world were larger.

## 5. Results

### *5.1. Simulation results*

Fig. 11a shows the performance of all planners on the detection of OOIs when the plants where roughly positioned in front of the robot. It was observed that the Semantic NBV planner significantly outperformed the other planners in terms of PCO at each viewing action and was able to detect 80% of the OOIs within nine viewing actions. After nine viewing actions, the Semantic NBV planner was able to efficiently search and detect 81.8% of the OOIs on average over 96 experiments, despite occlusions and uncertainty in plant positions. It perceived 13.3% more OOIs than the Volumetric NBV planner, 13.3% more than the Predefined Wide planner, 27.6% more than the Predefined Narrow planner, and 34.7% more than the Random planner. For targeted perception of plant parts, these results clearly show that viewing only the unexplored regions of the plant, as with the Volumetric NBV planner, is not efficient enough. It does not perform any better than a well-designed predefined strategy, such as the Predefined Wide planner. However, by adding knowledge of semantics, the performance improved significantly since the planner was now able to identify the relevant plant parts and prioritise them.

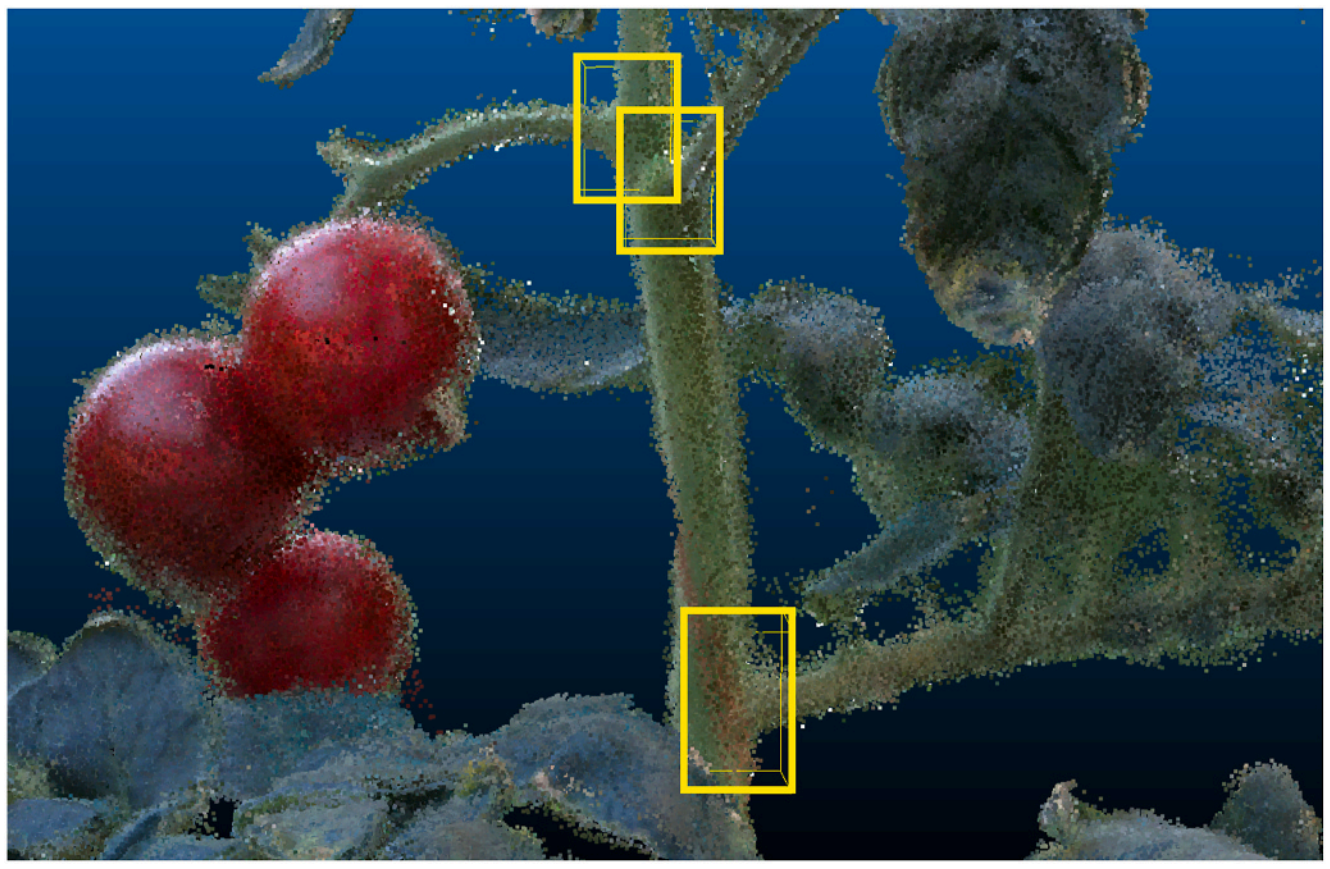

**Fig. 10.** A plant reconstructed using COLMAP, which was used as the ground truth for real-world experiments. The 3D positions of the objects-of-interest were annotated manually (marked in yellow).

The distribution density of the PCO after nine viewing actions averaged over all 96 experiments, shown in Fig. 11b, offers further insights on the performance of the planners. It was observed that the Semantic NBV planner's performance was more repeatable as indicated by the low spread of the density function and had a high accuracy with a median PCO of 87.8%, i.e. half the experiments had a PCO greater than this. The Volumetric NBV and Predefined Wide planners had a median PCO of 75.5% and 79.2% respectively, and their performances were less repeatable compared to the Semantic NBV planner as indicated by the wider spread of their PCO densities. The Predefined Narrow and Random planners were least repeatable as their PCO densities were spread over the entire range of PCO and their median PCOs were much lower at 70.6% and 50.0% respectively.

#### *5.1.1. Analysis of experimental and model parameters*

For brevity, only the three best performing planners from the previous experiment were analysed, namely Semantic NBV, Volumetric NBV and Predefined Wide planners. In the following analysis, the experiment presented in the previous section is referred as the *original experiment* and its results as the *original result*.

**Impact of removing uncertainty in plant position:** Fig. 12a shows the performance of the planners when the plants were placed in a fixed and known position in front of the robot. The Predefined Narrow planner is also shown in the plot because it detected the most OOIs after ten viewpoints. Compared to the original experiment with uncertainty in plant position (Fig. 11a), the Semantic NBV planner detected 8.7% more OOIs after nine viewpoints in this case, since it did not have to search for the plant before detecting the OOIs. The Predefined Narrow planner performed the best by detecting 96.0% of the OOIs after nine viewpoints. This indicates that the predefined planners in practice can be effective when the positions of the plants are known accurately. However, they are extremely sensitive to uncertainty in plant position and their performance can be unreliable, as seen from Fig. 11b. The Semantic and Volumetric NBV planners, on the other hand, are able to handle the uncertainty better by using the adaptive attention mechanism.

**Impact of removing uncertainty in OOI positions:** Fig. 12b shows the performance of the planners when the true OOI positions were known, i.e., all the OOI bounding boxes were provided at the beginning of the experiments. The Semantic NBV planner showed a small improvement and could detect 6.1% more OOIs after nine viewing actions compared to the original experiment. This implies that the Semantic NBV planner in the original experiment was able to explore and detect the OOIs quite reliably. Hence, its performance was close to the ideal case. On the other hand, the Volumetric NBV planner showed a larger improvement and could detect 11.7% more OOIs after nine viewing actions, indicating that it could not detect the OOIs reliably in the original experiment. The poor performance of the Volumetric NBV planner was likely due to the lack of an attention mechanism to focus on the OOIs. The performance of the Predefined Wide planner was independent of the OOI positions.

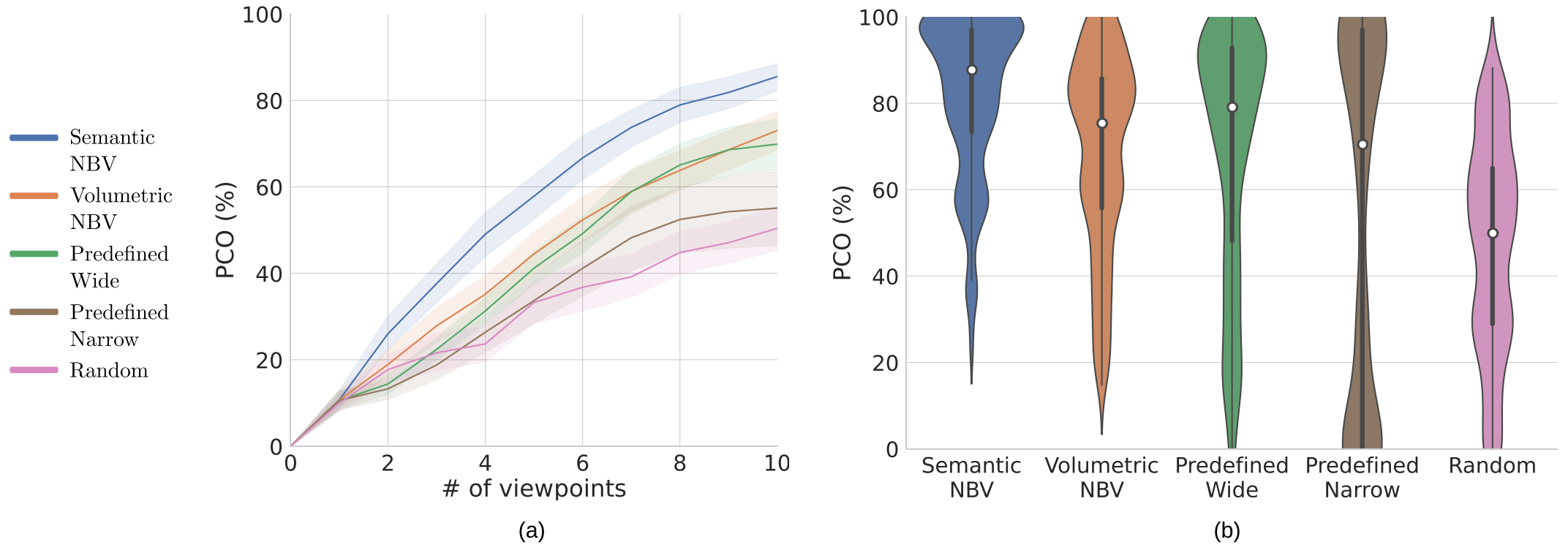

**Fig. 11.** Performance of the planners on the detection of all objects-of-interest (OOIs) when a tomato plant was placed 0.7m in front of the robot with an uncertainty of ±0.1 m in the x-axis and ±0.3 m in the y-axis. The percentage of correctly-detected objects (PCO) was determined for each viewing action. A detection was considered correct when at least 50% of the object was perceived (F1-score $\geq$ 0.5). (a) The mean PCO of the planners at each viewing action over 96 experiments. The error bands show the 95% confidence interval of the mean over all experiments. (b) The distribution density of PCO for the planners at the end of nine viewing actions over 96 experiments. The median PCO is shown by the white dot in the plot.

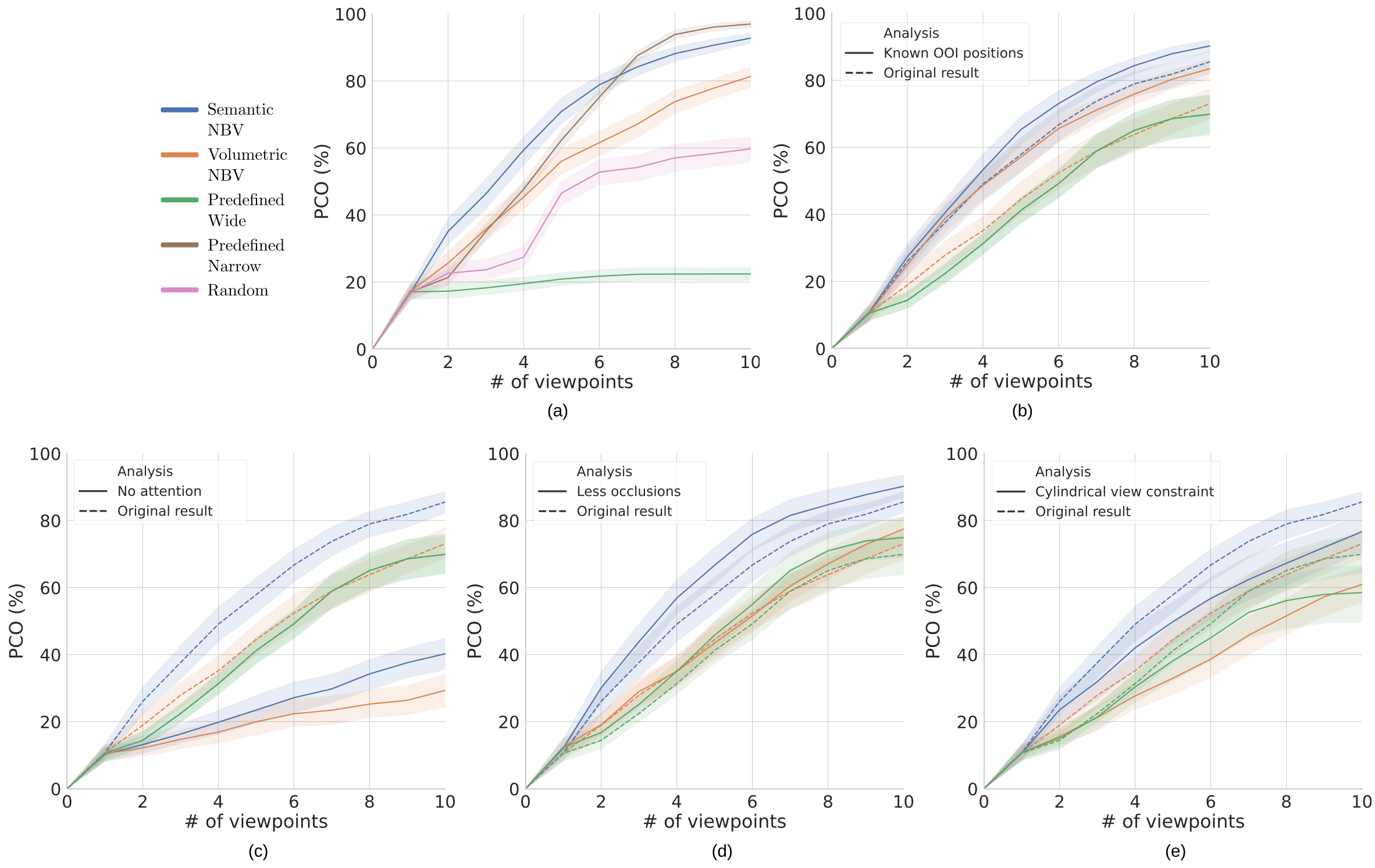

**Fig. 12.** Analysis of the performance of the planners based on various experimental and model parameters. The plots show the impact of (a) removing uncertainty in plant position, (b) removing uncertainty in OOI position, (c) removing attention, (d) reducing plant complexity or occlusion, and (e) changing the sampling constraint. The solid lines show the updated results based on the modified parameter and the dotted lines show the original results. The error bands show the 95% confidence interval of the mean over all 96 experiments. **Please note:** The solid and dotted lines of the Predefined Wide planner overlap each other in plots (b) and (c) since the performance of the planner was not influenced by these parameters.

**Impact of removing attention:** Fig. 12c shows the performance of the planners in the absence of an attention mechanism. The impact of removing attention was drastic on the Semantic and Volumetric NBV planners. Their performance dropped 44.3% and 42.2% respectively and they were unable to detect even half the OOIs after nine viewing actions. The drop in performance was observed because the planners had less guidance to perceive the relevant plant parts and spent time exploring irrelevant regions. This shows the significance of an attention mechanism in making NBV planning more efficient when performing a targeted search. Interestingly, the Semantic NBV planner outperforms the Volumetric NBV planner in this case too. This shows that the addition of semantic information, by itself, is helpful in guiding the planner to detect OOIs. When combined with an attention mechanism, the performance is further enhanced.

**Impact of reducing plant complexity:** Fig. 12d shows that the performance of all the planners improved when the plant complexity was reduced. The Semantic NBV planner could detect 5.8% more OOIs after nine viewing actions, while the Volumetric NBV planner could detect 4.2% more, and the Predefined Wide planner could detect 5.3% more. By removing the leaflets, the OOIs were less occluded and easier to detect from many viewpoints. The fact that only a small improvement in performance was observed indicates that all the three planners were able to handle the plant complexity and occlusions in the original experiment well. The overall results were still consistent with the original results, with the Semantic NBV planner outperforming the other planners by a significant margin.

**Sensitivity to sampling constraint:** Fig. 12e shows the performance of the planners when a cylindrical-sector constraint was used for view sampling (Section 4.2.2) instead of a planar-surface constraint. All the planners performed worse in this case. The Semantic NBV planner detected 9.8% less OOIs after nine viewing actions, while the Volumetric NBV planner detected 11.3% less and the Predefined Wide planner detected 10.6% less. Under the cylindrical-sector constraint, the viewpoints were forced to look towards the centre of the cylinder, which reduced the DoF from four to two. This potentially reduced the chances of finding good viewpoints and hence led to a drop in performance for all planners. The overall results were still consistent with the original results, with the Semantic NBV planner outperforming the other planners.

### 5.2. *Real-world results*

Fig. 13a shows the performance of the planners on the detection of petioles and peduncles of real-world tomato plants in a greenhouse. It was observed that the Semantic NBV planner outperformed the other planners in terms of PCO after five viewing actions and was able to detect 80% of the OOIs within seven viewing actions. After seven viewing actions, the Semantic NBV planner was able to search and detect 82.7% of the OOIs on average over 120 experiments, despite the real-world complexities such as natural variation and occlusion, natural illumination, sensor noise, and uncertainty in camera poses. The Semantic NBV planner detected 6.9% more OOIs than the Volumetric NBV planner, 15.8% more than the Predefined Wide planner, and 7.6% more than the Random planner. The Volumetric NBV planner did not perform any better than the Random planner in terms of the PCO, as it was attending to all unexplored regions of the plant and not attending specifically to the relevant plant parts. The Predefined Wide planner was the least efficient planner, which clearly indicates that using fixed or predefined viewpoints cannot handle occlusions well.

It was observed that the PCO of the Semantic NBV planner is similar to or slightly less than that of the Random planner in the initial viewpoints (Fig. 13a). This could potentially be due to false positives in the instance segmentation step (Section 3.1), in which Mask R–CNN sometimes mistook other plant parts as petioles or peduncles, as shown in Fig. 14b. False positives will cause the Semantic NBV planner to define a bounding box at a region that did not contain any OOI. Hence, the planner will be inefficient as it will unnecessarily spend viewpoints exploring an irrelevant region. However, in later viewpoints, the PCO of the Semantic NBV planner improved faster compared to the Volumetric NBV and Random planners, potentially because the falsely detected OOIs were sufficient explored, and the planner returned its focus to the true OOIs. The Semantic NBV planner had an average of one or two false positives per plant across all experiments (Fig. 14a). By reducing or handling these false positives, the performance of the Semantic NBV planner can potentially be improved further. Although the Volumetric NBV and the Random planners had a similar or higher number of false positives, these algorithms were not influenced by the false positives because they did not define bounding boxes on the OOIs.

The distribution density of the PCO after seven viewing actions averaged over all 120 experiments, shown in Fig. 13b, shows that the Semantic NBV planner's performance was highly repeatable as indicated by the low spread of the density function and had a high accuracy with a median PCO of 85.7%. The density functions of the Volumetric NBV and Random planners had a wider spread compared to the Semantic NBV planner and hence they were less repeatable. They also had a lower accuracy with a median PCO of 80.0% and 75.0% respectively. The Predefined Wide planner's performance was highly repeatable, but also the least accurate with a median PCO of 66.7%.

These results show that using semantic information about the relevant plant parts, the OOIs can be reliably and efficiently detect using fewer viewpoints. The results of the real-world experiments strongly support the results of the simulation experiments, indicating the advantage of using semantic information for NBV planning. Furthermore, the results are valid when using a real robot and camera setup under the complexities of real-world greenhouse environments.

## 6. Discussion

The simulation and real-world results show that the Semantic NBV planner is more efficient in detecting the OOIs compared the volumetric and predefined planners. By adding semantic information to an OctoMap and using it to define regions-of-interests, the planner chose viewpoints that paid attention to the OOIs and hence was able to detect them effectively. Compared to the literature in robotics, the results of this work are in line with Kay et al. (2021), who showed that distinguishing target objects from non-target objects using semantic information in NBV planning leads to faster reconstruction performance. The results are also in line with Zheng et al. (2019), who showed that the combined use of volumetric and semantic information leads to faster scene reconstruction and improved segmentation of objects. The main difference of this work from Kay et al. (2021) and Zheng et al. (2019) is the use of OOI clustering and an attention mechanism in combination with semantic information, which provided an object-level representation of the scene. This allowed the NBV planner to be aware of unexplored regions of each object. In the agricultural robotics literature, there were no prior works that directly incorporated semantic information for calculating the viewpoint utility. However, the works of Marangoz et al. (2022), Menon et al. (2023), Zaenker et al. (2023), and Zaenker et al. (2021) indirectly used semantic information to either identify the target regions or to estimate missing surfaces of the target objects. The results of this work are also in line with these works, which showed that using knowledge about the target region or the missing parts of the target objects can improve the performance of viewpoint planning for estimating the size, volume, or position of fruits. All these prior works, however, were evaluated only in simulation and were not quantitatively evaluated under real-world complexities.

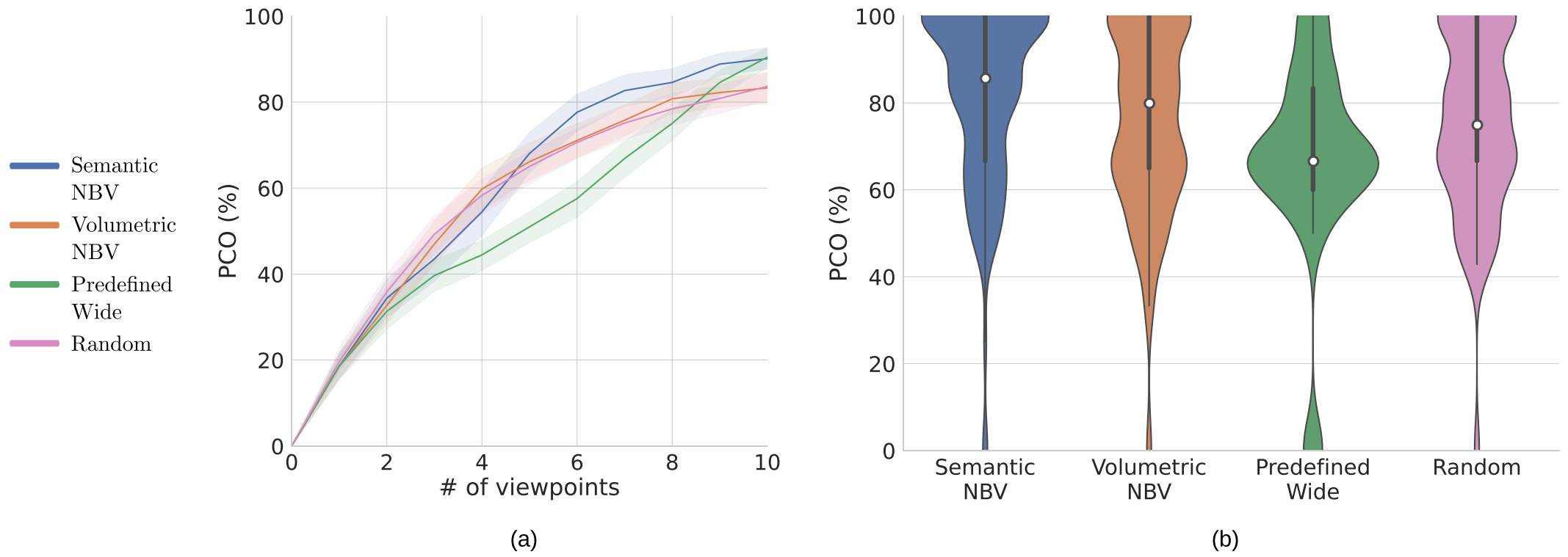

**Fig. 13.** Performance of the planners on the detection of petioles and peduncles of real tomato plants in a greenhouse. The percentage of correctly-detected objects (PCO) was determined for each viewing action. A detection was considered correct when at least 62.5% of the object was perceived (F1-score $\geq$ 0.625). (a) The mean PCO of the planners at each viewing action over 120 experiments. The error bands show the 95% confidence interval of the mean. (b) The distribution density of PCO for the planners at the end of seven viewing actions over 120 experiments. The median PCO is shown by the white dot.

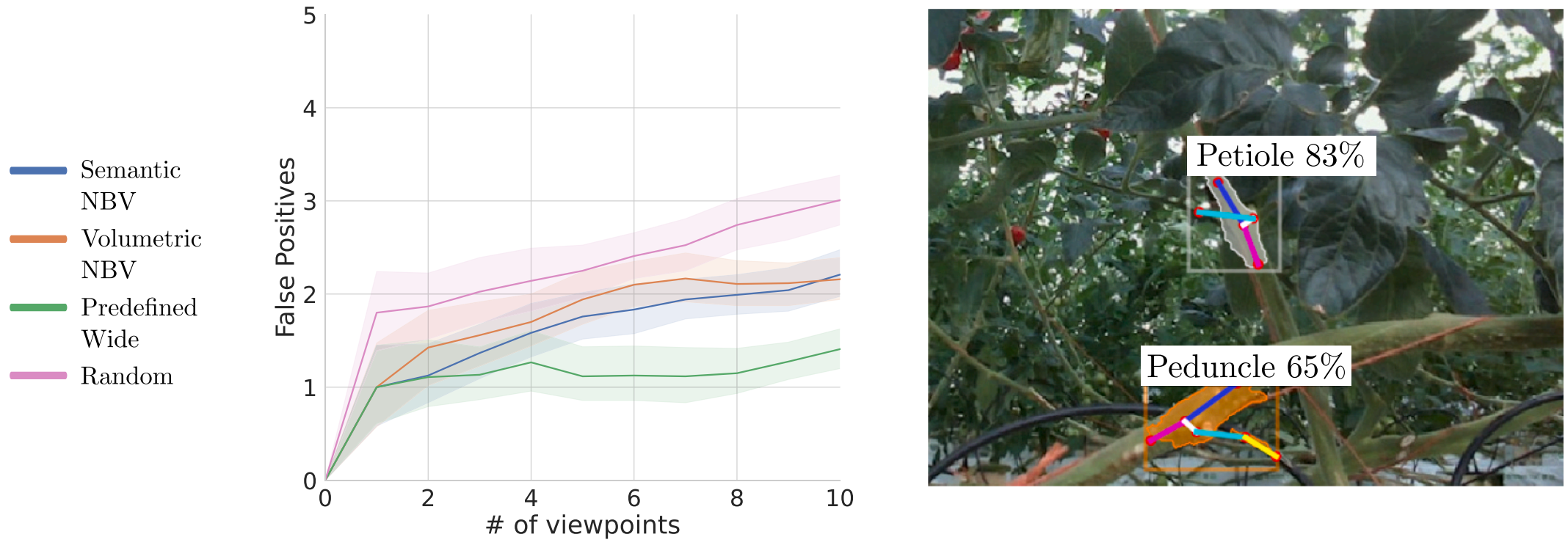

(a) (b)

**Fig. 14.** False positives from the instance segmentation algorithm influenced the performance of the Semantic NBV planner. (a) The mean number of false positives of the planners at each viewing action over 120 experiments. The error bands show the 95% confidence interval of the mean over all experiments. (b) Examples of two false positives in the tomato greenhouse. The top detection mistook a branch from a stem behind as a petiole, whereas the bottom detection mistook a cable passing across a stem in the front as a peduncle.

While the performance of the Semantic NBV planner was promising, there are certain aspects worth considering.

### 6.1. Influence of object-detection errors

The object-detection algorithm, Mask R–CNN, was able to detect most of the OOIs, but had some false positives and false negatives. False positives will cause the algorithm to explore irrelevant regions and make it inefficient. False negatives, on the other hand, will cause the algorithm to not pay attention to an OOI, leading to a poor perception performance. The impact of false positives and false negatives on the results was less severe since multiple viewpoints were used. Nevertheless, to improve the efficiency and accuracy of the planner, it is important to analyse the impact of these object-detection errors on the Semantic NBV planner and take measures to minimise them. Reducing false positives in semantic segmentation is difficult due to the complexity in greenhouses. There is always a chance for the instance segmentation network to observe something out-of-distribution and make a false detection. Future work could instead focus on effective strategies to remove false positives, for example, by verifying a detection from multiple viewpoints or estimating its uncertainty through repeated inferences.

### 6.2. Alternate ways to use semantic information

The confidence score from Mask R–CNN was used to define the semantic probabilities. This allowed us to calculate the entropy and choose viewpoints that minimised the entropy. While minimising the entropy is effective in improving the perception quality of the OOIs, it compromises slightly on the speed of exploring the scene for new OOIs. In the case of volumetric next-best-view planning, it has been shown by Isler et al. (2016) and Delmerico et al. (2018) that entropy-based approaches may not be the fastest way to explore the scene. Instead, using heuristics based on binary information about the presence or absence of an OOI could be faster in terms of exploration, while slightly compromising the perception quality. Future work could focus on comparing or even combining these two approaches to find a balance between exploration speed and perception quality.

A simple max-fusion method was used to update the semantic probabilities in the OctoMap. However, a more formal Bayesian update, similar to the update of the occupancy probabilities, might improve the accuracy of the semantic mapping. Bayesian updates consider prior probabilities and the likelihood of observations, potentially leading to a more robust and accurate representation of the environment. It could also help mitigate the influence of false positives, discussed in Section 5.2. Future research could explore the integration of Bayesian methods for updating semantic probabilities.

### 6.3. Influence of experimental and algorithmic parameters

The same initial viewpoint was used for all the planners to enable comparison between the experiments. However, the initial viewpoint could have an influence on the viewpoint planning. This was mitigated to an extent in the simulation experiments by rotating the plant along the z-axis. This was equivalent to changing the initial viewpoint around the plant. Also, the planners were free to move the camera anywhere on the planar surface defined in Section 4.2.1, meaning that the even if the initial viewpoint was observing the bottom of the plant, the next viewpoint cloud observe the top of the plant. Hence, we believe that the influence of the initial viewpoint on the performance of the planner was minimum.

A fixed number of candidate viewpoints were sampled at each step. Increasing the number of sampled candidates could potentially improve the performance of the planner. The influence of the number of candidates was studied in the work of Burusa et al. (2024) and no major difference in the reconstruction quality was found when 9, 27, or 45 candidates were sampled. We believe that this was because the viewpoints were constrained to face the plant and were well distributed across the viewing space. Hence, sampling as little as nine candidates could already provide a good performance. A similar influence of the number of candidates was expected in this work too because like in the work of Burusa et al. (2024), the viewpoint candidates were constrained and ensured that they were well distributed across the viewing space.

The OOI bounding boxes was predefined to a fixed size based on prior knowledge about the size of the plant parts. However, for applications where the size of the OOIs is not known or there are large variations in OOI sizes, it would be better to dynamically set the size of the OOI bounding boxes based on object detection. This could be achieved by, for example, using the bounding boxes from Mask R–CNN and the depth data to estimate the size of the detected objects. The segmented mask from Mask R–CNN and the depth data can also be used to complete the shape of the OOIs, for example, by fitting a superellipsoid (Lehnert et al., 2016). Dynamically adjusting the OOI sizes could potentially lead to better viewpoint planning since the planner will only spend an appropriate number of viewpoints on perceiving the OOIs.

### 6.4. Completeness of object perception

A threshold of 50% and 62.5% was used for the F1-score to consider an object as detected. For different applications, this threshold can vary depending on the desired perception accuracy. Although the sensitivity of the algorithm to the F1-score threshold was not studied in this paper, we believe that the Semantic NBV planner can still significantly outperform the other planners when the threshold is varied.

In the implementation, the F1-score threshold was only applied during evaluation. During execution, the Semantic NBV planner did not stop perceiving an object when the threshold was met but continued to perceive the object completely. Hence, the planner might have taken several extra viewpoints than necessary. If the level of object completeness required for the application is already know, the planner can be modified to plan viewpoints around an object only until the desired completeness is achieved.

### 6.5. Applicability in real-world conditions

The proposed method was evaluated using a dataset collected in a tomato greenhouse, which included real-world complexities such as natural variation and occlusion in tomato plants, natural illumination, sensor noise, and uncertainty in camera poses. The Semantic NBV planner was able to handle these conditions using a probabilistic 3D representation that merged semantic information from multiple viewpoints and kept track of the OOIs. Also, by using an adaptive attention mechanism, the planner was able to select the most informative viewpoints to perceive the OOIs.

The real world has a few other complexities that were not captured by the dataset developed in this work. In the dataset, the scene in the greenhouse was assumed to be static, but a tomato plant can move due to wind or contact by a human or the robot. The additional complexity introduced by such dynamics were not included in the experiments. These complexities lead to more uncertainty in perception of tomato plants, which only supports the need for active vision even more. Future work could develop methods to handle the dynamics in tomato greenhouses.

## 7. Conclusion

In this paper, a semantics-aware active-vision strategy for efficiently searching and detecting targeted plant parts, such as tomatoes, petioles, and peduncles was presented. A Semantic NBV planner, which added semantic information and an attention mechanism to conventional next-best-view planning was developed. Simulation and real-world experiments were performed to evaluate and gain insights into the behaviour of the planner, while ensuring repeatability and statistical significance. In simulation, the Semantic NBV planner detected 81.8% of the plant parts, about 13.3% more plant parts per plant compared to a Volumetric NBV planner, a commonly used active-vision strategy. Also, it detected about 13.3% and 27.6% plant parts more compared to two predefined planners and 34.7% plant parts more compared to a random planner. The Semantic NBV planner was able to search for the plant parts reliably with a median of 87.8% correctly detected objects in 96 experiments. It was also shown that the Semantic NBV planner was robust to uncertainty in plant and plant-part position, plant complexity, and different sampling constraints.

To the best of our knowledge, this work was the first to test the performance of a semantics-aware active vision strategy using a robotic setup in a real-world greenhouse environment. In real-world experiments, the Semantic NBV planner detected 82.7% of the plant parts, about 6.9% more plant parts per plant compared to a Volumetric NBV planer, 15.8% more than a predefined planner, and 7.6% more than a random planner. The real-world results strongly support the simulation results, despite the complexities of the real world, such as natural variation and occlusion, natural illumination, sensor noise, and uncertainty in camera poses.

The results of this work indicate that using an active-vision strategy with semantic information and an attention mechanism can effectively address the major challenge of occlusion in agro-food environments, thereby improving the accuracy and efficiency of perception. We hope it leads to significant improvements in robotic perception and greenhouse crop production in the future.

## Funding

This research was funded by the Netherlands Organisation for Scientific Research (NWO) project FlexCRAFT, grant number P17-01.

## CRediT authorship contribution statement

**Akshay K. Burusa:** Writing – original draft, Software, Methodology, Investigation, Data curation, Conceptualization. **Joost Scholten:** Software, Resources, Investigation, Data curation. **Xin Wang:** Writing – review & editing, Resources, Data curation. **David Rapado-Rincón:** Writing – review & editing, Resources, Data curation. **Eldert J. van Henten:** Writing – review & editing, Supervision, Funding acquisition, Conceptualization. **Gert Kootstra:** Writing – review & editing, Supervision, Funding acquisition, Conceptualization.

## Declaration of competing interest

The authors declare that they have no known competing financial interests or personal relationships that could have appeared to influence the work reported in this paper.

## Acknowledgements

We thank Thomas A.W. Versmissen for his help with annotating the real-world data and training the object-detection network. We thank the members of the FlexCRAFT project for engaging in fruitful discussions and providing valuable feedback to this work.